\documentclass[conference]{IEEEtran}
\usepackage{blindtext, graphicx}
\ifCLASSINFOpdf
\else
\fi

\begin{document}
%
\title{Computer-Assisted Text Analysis for Social Science: Topic Models and Beyond}

\author{\IEEEauthorblockN{Ryan Wesslen}
\IEEEauthorblockA{College of Computing and Informatics\\
University of North Carolina at Charlotte\\
Charlotte, North Carolina USA\\
Email: rwesslen@uncc.edu}
}


%


\maketitle

\begin{abstract}
Topic models are a family of statistical-based algorithms to summarize, explore and index large collections of text documents. After a decade of research led by computer scientists, topic models have spread to social science as a new generation of data-driven social scientists have searched for tools to explore large collections of unstructured text. Recently, social scientists have contributed to topic model literature with developments in causal inference and tools for handling the problem of multi-modality. In this paper, I provide a literature review on the evolution of topic modeling including extensions for document covariates, methods for evaluation and interpretation, and advances in interactive visualizations along with each aspect's relevance and application for social science research. 


\end{abstract}


%
\IEEEpeerreviewmaketitle

\section{Introduction}

Topic models are a framework of statistical-based algorithms used to identify and measure latent (hidden) topics within a corpus of text documents. Despite their wide use in computer science research \cite{blei2012probabilistic}, topic models have remained largely absent from the average social scientist's analytical toolkit. Historically, most social science text analysis has instead focused on either human coding or dictionary-based methods that are semi-automated but require high pre-analysis costs before implementation \cite{quinn2010analyze}. Moreover, this problem is magnified when considering the tremendous increase in the volume and variety of unstructured text documents for social scientists to study \cite{grimmer2013text}. To overcome this problem, social scientists have started to adopt computer-assisted text analysis techniques (like supervised learning and topic models) for their research with the goal of ``amplifying and augmenting'' social science analysis, not replacing it \cite{grimmer2013text}. 

Unlike computer scientists who typically use machine learning techniques for prediction, social scientists have found use in machine learning for the analysis of latent variables that could previously only be measured under untestable and consequential assumptions \cite{grimmer2015we}. Ultimately, the rise of computer-assisted text analysis tools in social science research is one of the major drivers of the emerging field of computational social science \cite{lazer2009life}, \cite{wallach2016computational}. The purpose of this paper is to survey the literature of one such computer-assisted text technique, topic models, and provide background on its importance to social science research. This overview leads itself to the newly created structural topic model (STM) that extends the general topic model framework to estimate causal effects within text documents \cite{stmss:roberts, openended:roberts}. 

In section 1, I review the evolution of topic models by introducing latent Dirichlet allocation (LDA) \cite{blei2003latent} and related seminal models. I also consider computational methods for topic models and discuss the extensions to the LDA-based framework to include document covariates within the model. In section 2, I explore tools for the application of topic models including methods for evaluation, measures of interpretation and visualization interfaces. In section 3, I discuss two major contributions by social scientists to the topic model literature: structural topic model (STM) and techniques to handle multi-modality. I provide examples of social science applications that use these techniques for texts like open-ended survey responses, political rhetoric, and social media. Last, I consider ongoing limitations in the methodology along with future opportunities for topic models within social science research in relation to computational social science and explainable artificial intelligence (XAI).

\section{Evolution of topic models}

Topic models identify and measure latent topics within a corpus of text documents.  Topic models are called generative models because they assume that observable data is generated by joint probability of variables that are interpreted to be topics. LDA is the workhorse topic model and is a Bayesian two-tiered mixture model that identifies word co-occurrence patterns, which are interpreted as topics. 

In this section, I summarize key properties of this framework by reviewing the evolution of topic models starting with its seminal models. Next, I review model extensions along with advances in computational methods in the model framework.

\subsection{LDA and Seminal Papers}

As a generalization, there are two approaches to computer-assisted text analysis: natural language processing (NLP) and statistical-based algorithms like topic models \cite{pLSI:hofmann}. Unlike NLP methods that tags parts-of-speech and grammatical structure, statistical-based models like topic models are largely based on the ``bag-of-words'' (BoW) assumption. In BoW models, collection of text documents are quantified into a document-term matrix (DTM) that counts the occurrence of each word (columns) for each document (rows). In the case of most topic models like LDA, the DTM is one of two model inputs (along with the number of topics). 

The BoW approach provides two key advantages (simplicity, statistical properties) at the expense of ignoring word order. For example, the BoW approach reduces the information contained in a collection of text documents into the word and document counts. An implication of word counts is that it ignores word order, which is opposite of NLP methods that parse language structure. Without accounting for word order, BoW methods perform poorly on micro-level problems like question-and-answer and others that require exact semantic meaning. However, for large collections of documents (sample size), the BoW assumption provides the theoretical foundation for a richer set of statistical methods (mixture models) by the assumption of exchangeability \cite{blei2003latent}. Ultimately, this advantage underpins statistical-based methods success in macro, document-level summarization problems for a large enough collection of inter-related documents.   

An early motivation for topic models was the goal of dimensionality reduction of the document-term matrix for large collections of documents or corpora. For example, Deerwester et al. (1990) presented one of the first predecessor models (latent semantic indexing, or LSI) by applying singular value decomposition (SVD), a linear algebra dimensionality reduction technique, to reduce the document-term matrix to latent factors. Their goal was to identify broad semantic (correlation) structure within the documents by removing noise from uninformative factors \cite{LSI:deerwester}. Later, Landauser and Dumais (1997) extended the LSI model to create the latent semantic analysis (LSA) model \cite{LSA:landauer}. Further, these methods could be improved by substituting the term-frequency inverse-document frequencies (TF-IDF) weightings in place of the raw term counts. However, Hofmann (2001) identified two major drawbacks to LSA. First, the approach lacked theoretical foundation as the method (SVD) relied on a Gaussian noise assumption that could not be justified for word counts (document-term matrix). Second, LSA could not account for polysemy, the multiple uses of words in different contexts \cite{pLSI:hofmann}. To address these problems, Hoffman (2001) introduced probabilistic latent semantic index (pLSI) model through the addition of a probabilistic (mixture) component to the LSA model by assuming each word is generated by a word probability distribution interpreted to be a ``topic''.\footnote{Alternatively, Ding et al. (2008) show a different but related methodology, Non-negative Matrix Factorization (NMF), is theoretically identical to pLSI as both approaches are maximizing the same objective function. The only difference between the two methods is that each approach differs in its inference method \cite{nmf:ding}.}

These seminal models set the stage for Blei et al. (2003) to extend these predecessor models to build the workhorse model LDA \cite{blei2003latent}. The key contribution of LDA was to extend Hofmann's pLSI model to include a second probability (mixture) component for the document-level, thus assuming documents are a mixture of topics. This addition yielded a two-tiered model, a core component of the typical topic model framework, in which observed words are assumed to be generated by the joint-probability of two mixtures. At the top, documents are a mixture of topics. At the bottom, topics are also a mixture of words. Each topic is defined as a unique distribution of words and yields the algorithm’s second output: the word-topic matrix. The word-topic matrix provides a conditional probability for every word (row) given each hidden topic (column). Using these probability distributions, a researcher can rank-order any word by each topic to determine what is the most common word the author(s) use when referring to each topic. \footnote{As will be discussed in Section 3, normally, the highest probability words conditioned on each topic serve to aid the researcher in the interpretation of each topic.} Similar to the idea of singular value decomposition (SVD) used in earlier topic models like latent semantic analysis (LSA), the probabilistic (mixture) nature of LDA acts similarly as dimensionality reduction process by reducing the information about each document from the large number of columns (words) to a much smaller number of columns (topics) \cite{dim:crain}. 

One consequence of the introduction of the mixture components within the LDA-based framework was the problem of ``intractability'' \cite{blei2012probabilistic}. In fact, like many other Bayesian methods, the introduction of the mixture components allowed, in theory, the measurement of the latent variable of topics but at the expense of the ability to measure precisely the optimal model because the exponentially large potential solutions of topic values. This problem lead to the question: what is the best approach to compute topic models?

\subsection{Computational Methods}

Following the introduction of LDA, a major theme in topic modeling literature was on computational methods for topic models given the problem of intractability of computing the evidence, or the marginal probability of observations \cite{blei2012probabilistic}, \cite{topic:blei}. Without an analytical solution, the goal of computational methods for topic model inference is to find the most (computational) efficient method that also best approximates the posterior. In general, there are two common approaches for topic model inference: sampling-based methods (e.g., MCMC/Gibbs Sampling) and variational inference. Sampling-based algorithms simulate samples of the posterior to approximate the true posterior. Gibbs sampling, the most common sampling method, was introduced for topic model inference by Griffiths and Steyvers (2004) and uses a Markov Chain to estimate a sequence of dependent random variables that asymptotically serves as the posterior distribution \cite{scitopic:griffiths}. Ultimately, sampling-based methods like Gibbs sampling have the advantages that they are (1) theoretically backed, (2) unbiased, and (3) computationally convenient to implement \cite{scitopic:griffiths}. The downside of this approach is that it can be very slow for large inputs (number of documents, words or topics).

\begin{figure}[!t]
\centering
\includegraphics[width=3.5in]{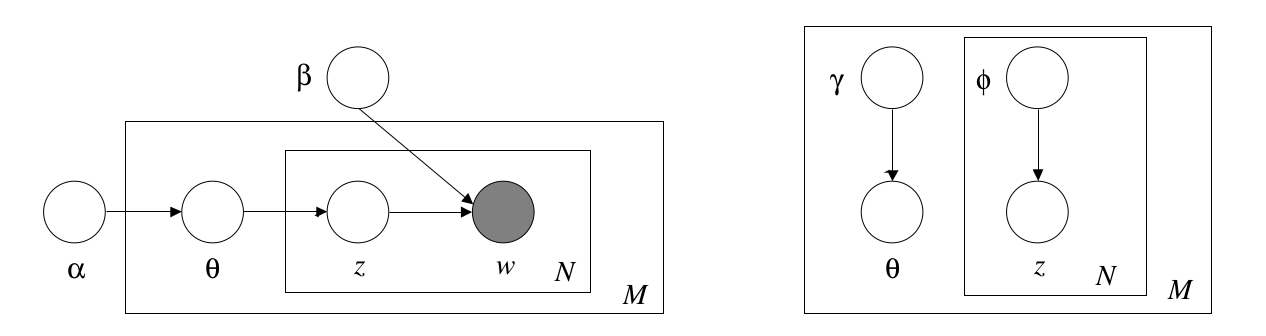}
\caption{(Left) Graphical model for latent Dirichlet allocation (LDA) and (Right) a graphical model of the variational distribution used for variational inference for LDA from Blei et al. (2003)}
\label{fig_lda}
\end{figure}

As an alternative to approximating the posterior through sampling, variational inference methods transform the problem into an optimization problem. In this context, the goal is to find families of distributions over the hidden variables (topics) that most closely estimates the actual posterior \cite{blei2012probabilistic}. In other words, variational inference attempts to most tightly estimate the posterior with a simpler distribution that includes free variational parameters used as the optimization arguments \cite{topic:blei}. Blei et al. (2003) introduce variational methods for topic model inference by using an Expectation Maximization (EM) algorithm. Figure \ref{fig_lda} provides the two graphical models, the left the theoretical model for LDA and the right the simplified model used in the variational inference algorithm. To use variational inference for LDA, the model is simplified by removing the edges between $\theta$, \textbf{w}, and \textbf{z} variables and introducing the free variational parameters $\gamma$ and $\phi$ \cite{blei2003latent}. The optimization problem uses Kullback-Leibler (KL) divergence to best minimize the estimated posterior to the true posterior. The EM algorithm is used as a two step process in which the variational distribution is estimated with estimated parameters and then the new variational parameters are chosen through the variational inference optimization problem. The process is repeated until a convergence threshold is met. Hoffman et al. (2010) extended the variational inference to introduce a faster online batch algorithm that can be used to massively scale LDA for very large corpora or streaming data \cite{2010:hoffman}. 

Ultimately, given that sampling-based or variational inference methods are estimates and never ever exact solutions, neither method is perfect and the decision of each depends on the trades off of speed, complexity, accuracy and simplicity required for the problem at hand \cite{topic:blei}. In the next section, I'll introduce how the basic LDA framework has been generalized to include document metadata variables within the algorithm which thus require modifications to the computational methods as well. Later in Section 4,  I'll discuss how such metadata variable extensions has an impact on the computational methods possible when introducing the structural topic model (STM).

\subsection{Model Extensions for Document Metadata}

A second theme in topic modeling literature deals with the inclusion of metadata variables into the model. One of the first examples of such a model is the author-topic model proposed by Rosen-Zvi et al. (2004). The original motivation was the explicit point that who the author is will have a direct impact on what topics are discussed in a publication (e.g., a biologist will more likely write about topics in biology than sociology or politics). In contrast, LDA does not take into account any document variables (like author) and thus fails to incorporate author into the model. Using only LDA, the only way to analyze the impact of author ``post-hoc'' was by comparing how the model outputs (topics) compare relative to author. A noted downside of this approach is that the model would likely be less effective as its omitting a known variable that affects the topic proportions. To address this problem, Rosen-Zvi et al. (2004) introduced the author-topic model to incorporate the author attribute by modifying LDA's assumption that author, not documents, are a multinomial distribution over the topics \cite{author:rosen}. Soon after, many metadata topic model extensions were created for a variety of metadata attributes like time (dynamic topic model \cite{dtm:blei}), geography (geographical topic model \cite{gtm:eisenstein}), and emotion (emotion topic model \cite{etm:rao}).

Given the large collection of metadata topic model extensions, Mimno and McCallum (2009) categorized metadata extension models into two groups: down-stream and up-stream models. The key difference between each approach is on the role the metadata variables take in the process to generate the text. For instance, in the down-stream approach, the metadata (like the text itself) is assumed to be generated by the hidden topics. In this approach, topics are word distributions as well as distributions over the metadata variables. Figure 2 provides an example plate notation for a down-stream model. In this figure, the metadata variables (m) and the words in the text (w) are conditionally generated by the hidden topics (z). The most common example of this approach is the supervised latent Dirichlet allocation (sLDA) model \cite{supervised:blei}. 

\begin{figure}[!t]
\centering
\includegraphics[width=2.5in]{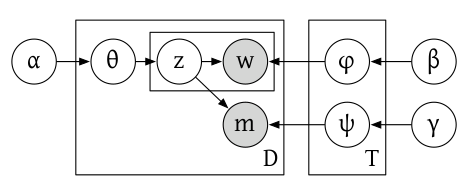}
\caption{Example of an down-stream model extension from Mimno and McCallum (2008)}
\label{fig_down}
\end{figure}

Whereas in the up-stream approach, the algorithm is conditioned on the metadata covariates such that the document-topic distributions are mixtures of the covariate-specific distribution \cite{dmr:mimno}. The classic example of this approach is the author-topic model or dynamic topic model. Essentially, the up-stream models ``learn an assignment of the words in each document to one of a set of entities'' \cite{dmr:mimno}.

\begin{figure}[!t]
\centering
\includegraphics[width=2.5in]{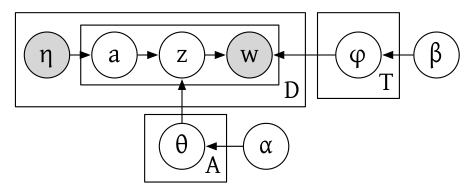}
\caption{Example of an up-stream model extension from Mimno and McCallum (2008)}
\label{fig_up}
\end{figure}

In addition to incorporating metadata into the model, other topic model extensions include additional information like word context, graphical (network) relationships and hierarchical topic structure. For example, the Hidden Markov Model (HMM) extended the normal BoW assumption to facilitate a consideration of word context into the topic model framework \cite{hmm:gruber}. In the case of network data, the link-topic model \cite{topiclink:liu} and the relational topic model (RTM) \cite{relational:chang} are two additional topic models that incorporate relational based information into the model for further analysis. Finally, Teh et al. (2006) introduce a generalized hierarchical structure to the topics that alleviates the problem of the number of topics and allows analysis from different topic levels \cite{hdp:teh}. Nevertheless, one of the main advantages of topic models is its flexibility to incorporate multiple different computational methods and alternative specifications. In the next section, I'll move on to more practical considerations of how to use and apply topic models before introducing social science applications in Section 4.

\section{Tools for Model Application: Interpretation, Model Selection and Visualization}

In this section, I review research that has focused on pre-processing, interpretation, evaluation and analysis of topic models. While not necessarily changing the underlying model framework like discussed in Section 2, this research is critical for the implementation of topic models and their application for social science research.

\subsection{Pre-processing}

A critical step to analyze text is the process of quantifying text. This process, called pre-processing, is a series of decisions a researcher makes to clean and normalize text with the goal of removing potential noise to maximize her analysis on underlying signal. While many researchers overlook these steps, relying instead of default rules without questioning their merits, recent research has found that text analysis methods like topic modeling are susceptible to potential ``forking paths'' that leave results subject to initial coding decisions \cite{stat:gelman, text:denny, preprocess:schofield}. Moreover, there is no set rules for what pre-processing steps are necessary and the need is defined by the quality, quantity and style of the underlying text. 









\subsection{Model Selection: Prediction-Interpretability Trade-off}

The question of how to select and validate topic model specifications (e.g. number of topics, model framework, etc.) depends on the researcher's objective. In predictive modeling, a researcher's goal is to build a model that can best predict out-of-sample or future documents using log-likelihood (perplexity) measures. Whereas for researchers whose goal is exploration and knowledge discovery, human judgment (e.g. interpretable topics) may take precedence over holdout prediction. 

Initially in the topic model literature, prediction accuracy was the key goal of model evaluation. Wallach et al. (2009a) outlined different evaluation methods on LDA using Gibbs sampling. They consider two approaches to evaluating topic models: maximizing the held-out documents likelihood (perplexity) and document completion in which long documents are trained on part of the document and evaluated on the model's ability to correctly ``complete'' the document. They find that methods like harmonic mean, importance sampling and document completion methods are inaccurate and may distort the relative advantage of one model versus another model. Instead, they recommend either the Chib-style estimator or the ``left-to-right'' algorithm as more accurate evaluation methods \cite{eval:wallach}.\footnote{Two limitations to Wallach et al. (2009a) is that these results were only tested using Gibbs sampling-based inference and on vanilla LDA.} 

However, Chang el al. (2009) explored the trade-off between prediction and interpretability. Through the word intrusion tasks, they found the counter-intuitive result that highly predictive topics tend to be negatively correlated with interpretability. Semantic coherence was introduced by Mimno et al. (2011) as a measure for how internally consistent words are within topics. On the other hand, exclusivity is a measure to identify words that have high probabilities for only a few topics rather than many topics. Roberts et al. argue that a ``topic that is both cohesive and exclusive is more likely to be semantically useful'' \cite{openended:roberts}. 

Finally, other researchers have studied the effectiveness of topic modeling while controlling for other key factors (e.g. hyperparameters, number of topics, document sample size, document length). Wallach et al. (2009b) explored the effect of relaxing LDA's prior distribution assumptions, including using non-symmetric Dirichlet parameters for the document-topic or word-topic matrices. They find that asymmetric priors on the document-topic distribution can provide substantial advantages while asymmetric priors do not provide much benefit to the word-topic distribution. Taddy (2012) explored estimation methods for choosing the optimal number of topics via block-diagonal approximations to the information matrix and goodness-of-fit analysis for likelihood-based model selection \cite{eval:taddy}.

In a more systematic review of topic model performance, Tang et al. (2014) analyzed LDA performance controlling for four limiting factors: document length, number of documents, and the two prior distribution hyperparameters. Considering two simulated and three real datasets (Wikipedia, New York Times, and Twitter), they make five recommendations. First, they argue that a sufficient number of documents is the most important factor to ensure accurate inference. For example, they find LDA is more difficult when running on a small sample (e.g. less than a thousand documents). Second, they find the length of the document matters as well. This matters in social media data like Twitter messages in which all messages are less than 140 characters. As an alternative, they cite alternatives like aggregating messages to transform the documents to a user-level to expand the size of documents \cite{twitter:hong}. Third, they find that collections with too many topics lend statistical inference methods to be inefficient. Fourth, they find that LDA performance is affected by how well-separated the underlying topics are relative to a Euclidean measure. Last, they find that the variability of hyperparameters is important depending on the number of topics within documents. For example, they recommend using a lower alpha (Dirichlet parameter) when documents have few topics, whereas using a high alpha for documents with many topics.

\subsection{Visualizations}

In this section, I review semi-automation methods of analyzing topic models through visualizations. Hu et al. (2013) argue that topic models suffer from an interpretation problem that requires the need for interactive system for end users. Topic results are never perfect and often include bad topics that do not perfectly align with an end-user's judgment and intuition (similar to the argument by Chang et al. 2009). To address this problem, they argue for systems that allow the end user to annotate the model results and incorporate feedback into the model's output. Moreover, they identify social science (in addition to digital humanities and information studies) as a discipline that would greatly benefit from interactive systems when implementing topic models. They emphasize social science as such a field because of its ``take it or leave it'' problem because many social scientists ``have extensive domain knowledge but lack the machine learning expertise to modify topic model algorithms'' \cite{2013:hu}. A shortcoming of their argument is they omit the role that visualizations can play within such interactive systems. Further, they fail to recognize the body of research by the visualization community that has extended and proposed many different applications for visualizations to fit into such interactive systems.

Dou and Liu (2016) argue that visual interfaces allow decision makers to explore and analyze the model results. This point is more important when considering the application of topic models for non-computer scientists (like most social scientists) who may not have the programming or computational training to run the algorithms on their own. A major motivation for the use of visualizations for analyzing topic models is that the output is too large for a researcher to absorb manually \cite{survey:dou}. For example, LDA's output is two large datasets (the word-topic and the document-topic matrices), the size of both are proportional to the number of documents, terms and topics. Therefore, the larger the corpus, the larger the output and the more difficult it is for a researcher to analyze the results. 

\begin{figure}[ht]
\centering
\includegraphics[width=3.5in]{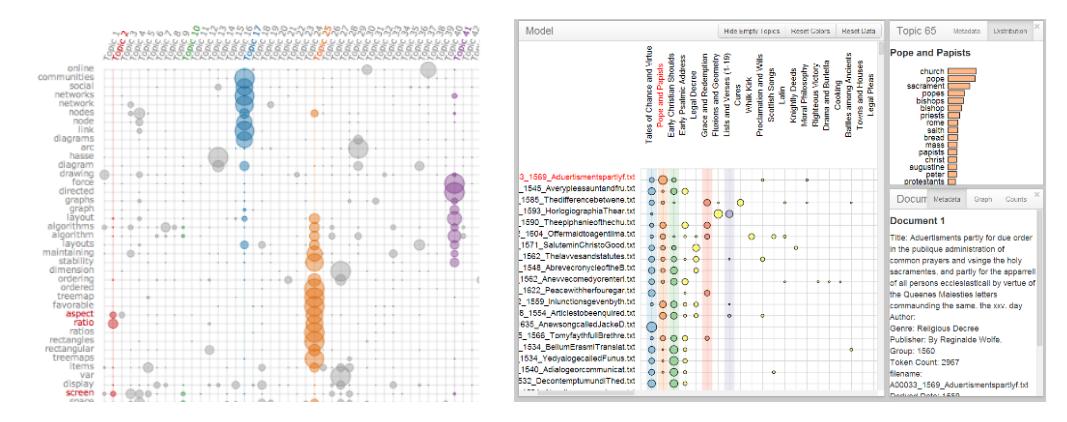}
\caption{Topic-oriented visualizations including (Left)the Termite model from Chuang et al. (2012) and (Right) the Serendip model from Alexander et al. (2014)}
\label{oriented}
\end{figure}

In general, there are two common approaches to visualizing topic models: topic-oriented and time-oriented \cite{survey:dou}. Each approach differs based on which is the most important element of interest. In topic-oriented visualizations, the focus is on the relationship between either the words and topics (word-topic matrix) or the document and topics (document-topic matrix). Such approaches focus on the task of document summarization, information retrieval and relationships between documents. Common examples of these approaches include matrix representations like Termite \cite{termite:chuang} and Serendip \cite{serendip:alexander} (see Figure \ref{oriented}) as well as parallel coordinates visualizations as in ParallelTopics \cite{parallel:dou}. Chuang et al. (2012) provide a general design framework for topic-oriented interactive visual systems based on how an analyst makes inference on the topics (interpretation) and the actual and perceived accuracy of the analyst's inference (trust) \cite{trust:chuang}. Other interfaces like HierarchicalTopics have generalized the model and facilitated interfaces that focus on a hierarchical structure within the topics that can aid in drill down on multiple levels for document summarization \cite{hier:dou}, \cite{hierar:cui}. Moreover, new research has used (network) graphs to represent the correlations between topics, especially when coupled with models with a more flexible correlation structure like CTM \cite{pano:wang}.

On the other hand, documents that are time-oriented (e.g. Twitter messages or news articles) can be aided with time-dependent visualizations that can aid in exploring the trend, evolution, lead-lag effect and event-detection relative to the topics. TIARA is an interface created to visualize topical trends by using an enhanced stacked graph \cite{tiara:wei}, \cite{tiara:liu}. Similarly, TextFlow was introduced with the goal of exploring the evolution of topics including identifying how topics merge and split over time \cite{textflow:cui}. Further, TextPioneer is a visual interface that introduces the problem of “lead-lag” relationships in exploring topic results \cite{textpioneer:liu}. Lead-lag is the problem in which a researcher needs to understand the relationship between two corpora, especially in the case when one corpus (e.g. social media) may lead the information that may then appear in another corpus (e.g. news articles that recount information spread through social media). Last, another time-oriented component that has been explored in visual interfaces is event-detection and analysis. For example, LeadLine is a visual analysis system used to identify and explore events by detecting the most common words (topics) used in short, discrete bursts \cite{leadline:dou}.

\begin{figure}[ht]
\centering
\includegraphics[width=3.5in]{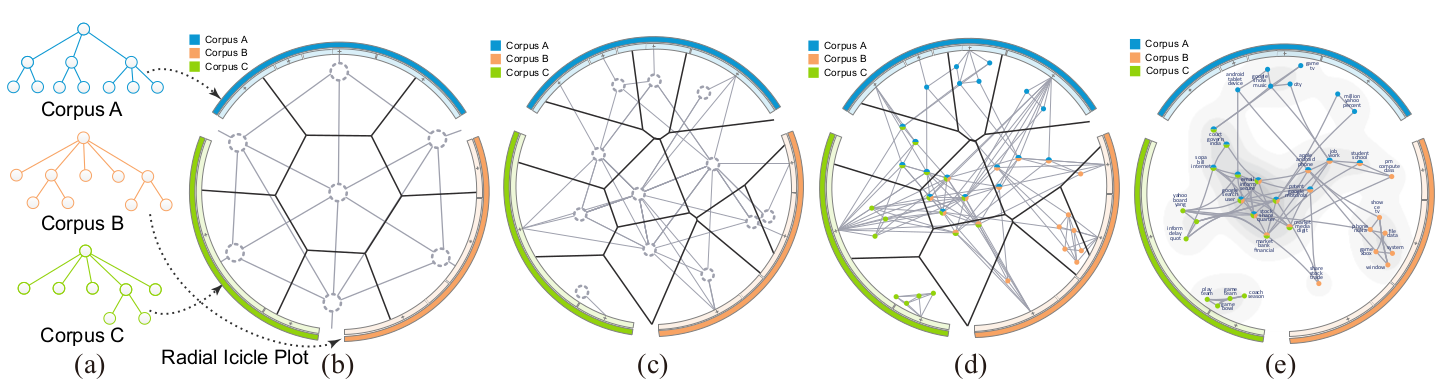}
\caption{The integration of multiple corpora within TopicPanorama from Liu et al. (2016)}
\label{topicpanorama}
\end{figure}

Last, another major consideration in the use of visual interfaces for topic models includes the type of data used within the model. The simplest approach is when homogeneous data is used, and thus the focus is on a single corpus and the topics that stem from the corpus. However, much deeper insight can be found with the inclusion of heterogeneous data sources that append document metadata to the corpus. One example of a recent approach to combine such sources include the TopicPanorama interface that combines text from multiple data sources (e.g. news articles and Twitter messages) and provides a network graph to link across these sources (see Figure \ref{topicpanorama}. Further, another avenue of topic analysis includes the impact of analyzing topics within streaming data sources like Twitter or other social media platforms \cite{textstream:liu}.

\begin{figure*}[t!]
\centering
\includegraphics[width=6.5in]{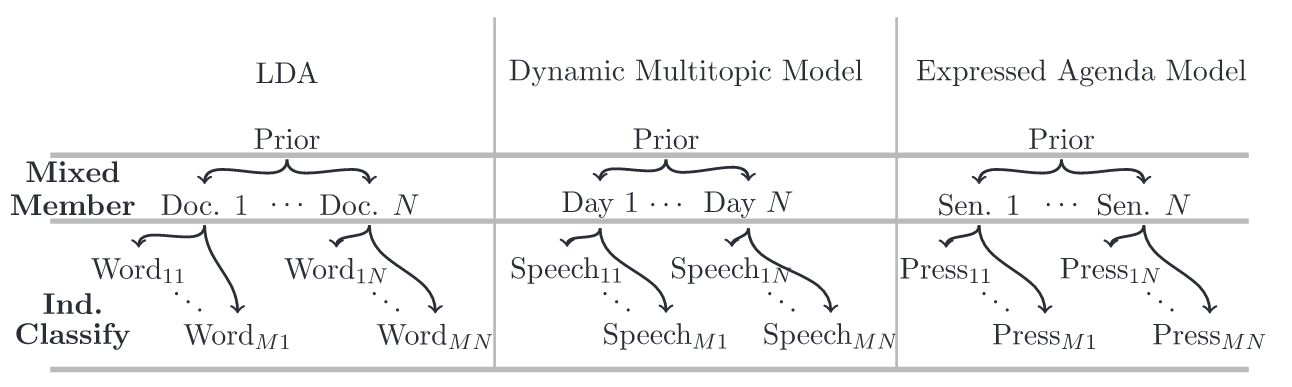}
\caption{Comparison of LDA, Dynamic Multitopic Model and Expressed Agenda Model from Grimmer and Stewart (2013)}
\label{fig_poli}
\end{figure*}

\subsection{LDA-based Topic Model Applications in Social Sciences}

One of the first applications of topic models within social science literature comes from Quinn et al. (2010). This paper laid the foundation for social science (mainly political science) application of topic models in three ways. First, the paper directly compares topic model relative to other text analysis methods (reading, human coding, dictionary-based and supervised learning), comparing and contrasting the costs and benefits of each method.\footnote{As mentioned in the introduction of this paper, this cost-benefit comparison justified the benefit of computer-assisted text analysis tools like topic models relative traditional text analysis research like human coding and dictionary-based methods \cite{quinn2010analyze}.} Second, Quinn et al. set out a list of five criterion-based concepts (goals) based on more traditional social science content analysis to be used for model evaluation that, they argue, should be considered when applying topic models.\footnote{The five criterion are: semantic validity, convergent construct validity, discriminant construct validity, predictive validity, and hypothesis validity. Semantic validity is the coherent meaning from interpreting the word-topic probabilities. Convergent construct validity is the extent that the results align to existing measures or benchmarks of known truths. Discriminant construct validity measures how the results depart from the same existing benchmarks. Predictive validity is the ability that the results can correctly predict external events. Last, hypothesis validity is the extent to which the results can be effectively used to test hypotheses \cite{quinn2010analyze}.} Last, the authors modified LDA for their analysis by using a continuous time model that limited the number of topics for each speech to only one (as opposed to the mixture assumed in LDA). The model is named the Dynamic Multitopic Model and is represented in Figure 5. Unlike LDA, the model is a single-membership mixture model in which each speech was assigned to a unique topic, and each day was assumed to be a mixture of speeches. As shown in Figure \ref{fig_poli}, this can be analogous to how LDA assigns each word to a topic, then assumes each document is a mixture of those topics. This is important as it was one of the first social science-based topic model modifications created to test a theoretically driven hypothesis for social science research.

Similarly, Grimmer (2010) created a modified topic model, the Expressed Agenda model, to analyze political rhetoric by individual senators through their press releases \cite{2010:grimmer}. Also shown in Figure 6, the Expressed Agenda model modified the LDA framework into a single membership mixture model and assumed each press release was assigned to a single topic, while each senator's press releases corpus were assumed to be a mixture of those topics. Like the Dynamic Multitopic Model, the units of measurement (e.g. topic assignment and document-level) were modified from standard LDA to allow the author to test a political theory on how a senator divides his or her attention across the multiple political issues as represented by their explicit press releases \cite{2010:grimmer}.

\begin{figure}[h!]
\centering
\includegraphics[width=3.25in]{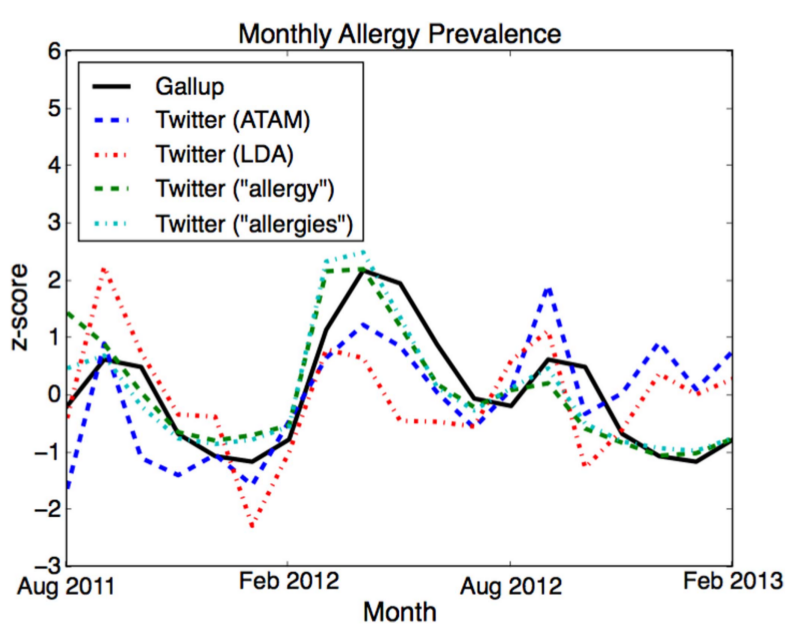}
\caption{Monthly allergy prevalence for Tweets using four methods and Gallup survey results from Paul and Dredze (2014)}
\label{fig_health}
\end{figure}

Other social scientists have considered using LDA to analyze text from social media (in particular Twitter). One novel example comes from health informatics by Paul and Dredze (2014). Using Twitter messages, the authors modify LDA to create the Ailment Topic Aspect Model (ATAM) with the goal of identifying health related keywords automatically. The authors discover 13 interpretable topics (e.g., seasonal flu, allergies, exercise, and obesity) that correlate significantly with U.S. geographic survey data \cite{2014:paul}. Figure \ref{fig_health} provides an except of their findings by comparing the normalized frequency (z-score) trends of external survey results on allergies and four approaches to identify allergy-related Tweets (by keywords ``allergy'' and ``allergies'', LDA and ATAM). The key methodological contribution of this model was the inclusion of external ``background words'' that represent health aspects that the model will identify as topics that describe the aspect. Further, the authors explore their output topics relative to document-level metadata regarding the time and geography of the Tweet. A limitation of their approach is that, while the results correlate with external (Gallup) benchmarks, their work does not demonstrate with statistical significance the relationship between Twitter topics and its effect on public awareness (via the Gallup results). At its core, this approach lacks a causal inference mechanism to explain the significance of the discussions on Twitter.

\begin{figure}[h!]
\centering
\includegraphics[width=3.5in]{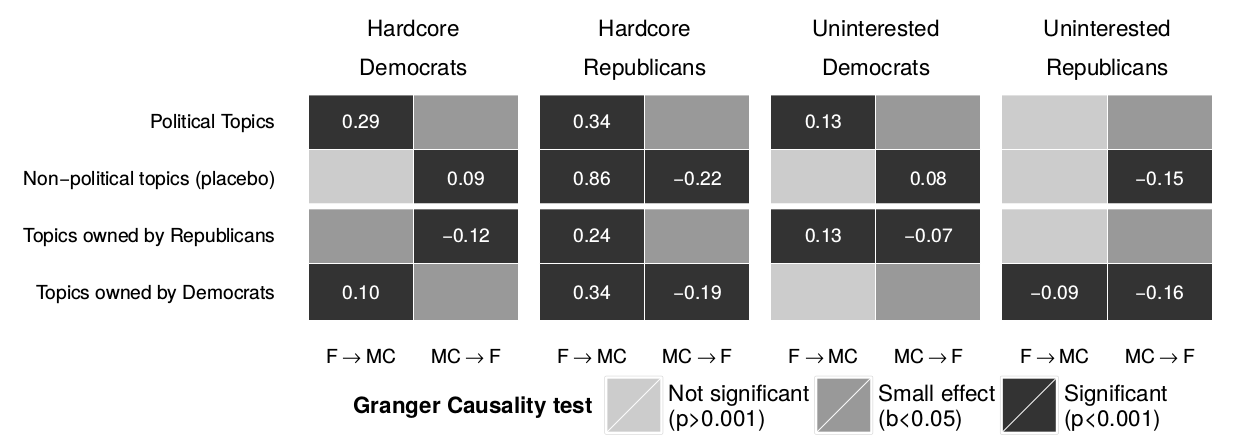}
\caption{Differences in the estimated responsiveness of different constituent follower groups using F-tests from Barbera et al. (2014)}
\label{fig_bar}
\end{figure}

One application of LDA on Twitter that attempts to provide a causal inference mechanism comes from political science by Barbera et al. (2014). In their paper, the authors analyzed the responsiveness to members of the United States Congress to constituent conversations on Twitter. In order to categorize the topics, they used LDA on the tweets of both constituents and the members of Congress. Next, they measured the variance between the topics over time to estimate whether members of Congress lead or follow their constituents on political issues by employing Granger-causality testing after running LDA \cite{2014:barbera}. The key contribution of this paper was the use of a temporal causal framework (Granger causality) along with LDA to provide a higher level statistical significance for the effect of covariates (author and time). To analyze this result, they divided the members of Congress and their constituents into six groups, three groups per political party (Democrat and Republican): members of Congress, ``hardcore'' constituents and ``uninterested'' constituents. Also, they categorized the topics from LDA into four categories: Democrat-owned, Republican-owned, non-political and political topics. Then they ran 24 ``post-hoc'' regressions (four topics x six groups) using five day lags of topic proportions for each of the six groups as the 50 independent variables. Figure \ref{fig_bar} provides standard F-test results by topic (rows), columns (group) and the relationship.\footnote{The ``F $\rightarrow$ MC'' column represents F-test for measuring the followers impact on the member of Congress' topic proportions while the ``MC $\rightarrow$ F'' column is the opposite (i.e., a member of Congress' impact on his/her followers' topic proportions)}. They find that members of Congress are responsive to their constituents, especially prominent issues by their ``hardcore'' constituents. However, they find little evidence of that members of Congress have influence on what topics their constituents discuss publicly \cite{2014:barbera}.

While being a novel contribution to estimating causal effects with LDA, their approach has two limiting factors. First, to analyze author and time, they did not directly model. Instead, employing the ``aggregation'' approach suggested by \cite{twitter:hong}, they combined Tweets by author and day to modify the definition of a document from a tweet to the collection of all tweets in a day by each author. By doing aggregation, this enables them to control for the document covariates of time and author but still employing LDA that does not have a mechanism to directly control for these covariates. Second, they use regression ``post-hoc'' and not within the generative topic model itself. As noted in the appendix of Roberts et al. (2014), the problem with this approach is the measurement of uncertainty that can lead to spurious results \cite{openended:roberts}.\footnote{Roberts et al. (2014) call this approach two-stage in which in LDA is run (stage 1) followed by running a regression for topic proportions conditioned on a covariate of interest (stage 2).} 



A general theme of these applications represent the flexibility of how LDA-based framework can be modified to address a unique theoretical question for a specific document-level covariate (e.g. time, author, geography). However, Grimmer and Stewart (2013) recognized that requiring all social scientists to ``tune each model to their task is a daunting task''. Further, each provides the motivation for deriving a causal inference component within the LDA framework by asking questions involving document covariates that facilitate hypothesis testing. In response to these problems, Roberts et al. (2013) introduced structural topic model (STM) as a general causal inference framework for hypothesis testing for document covariates.

\section{Structural Topic Model \& Multi-Modality}

In this section, I review literature by social scientists to reconcile two major problems with the standard topic model framework: the lack of causal inference and multi-modality.

The first issue I address is that standard (down-stream) metadata topic models (e.g. Author-Topic, Dynamic, etc.) make point but not standard error estimates to facilitate statistical hypothesis testing.  The next problem is that computational methods for topic model inference, as it is an NP-hard problem \cite{np:sontag}, \cite{localmodes:roberts}, can provide local optima but cannot guarantee global optima, which is termed multi-modality. This problem threatens the stability of a topic model output and can lead researchers to question whether they ``did not stumble across the result completely by chance'' \cite{localmodes:roberts}. 

\subsection{STM Model}

The first major critique of topic models is that its output provides point estimate of word or topic probabilities without confidence intervals that facilitate statistical hypothesis testing. This problem is especially important in down-stream extensions that include metadata that affect the topic proportions. For example, researchers could estimate that a topic proportion is different for various covariate levels, but without statistical confidence provided by standard error estimates. This problem is directly contrary to the tradition of causal inference that employs statistical confidence within the social sciences to determine causation (for example \cite{2011:greene}). In an effort to reconcile this problem, Roberts et al. (2013) and Roberts et al. (2014) introduced the structural topic model (STM) by incorporating a generalized linear model (GLM) framework for document metadata by extending elements of three previous topic model extensions: CTM, DMR and SAGE. 

Let's first start with CTM. Blei and Lafferty (2007) introduce the correlated topic model (CTM) to provide more realism to the original LDA model which incorporates a more flexible correlation structure than the independence assumption assumed in LDA. The model replaces the Dirichlet assumption for topic proportions as used in LDA with a logistic normal distribution. The main advantage of the CTM versus the LDA is improved predictive power: ``A Dirichlet-based model will predict items based on the latent topics that the observations suggest, but the CTM will predict items associated with additional topics that are correlated with the conditionally probable topics'' \cite{ctm:blei}. 

However, like many other topic model extensions, relaxation of model assumptions comes at the expense of model complexity and even intractability for existing methods. In this case, simulation techniques like Gibbs sampling are no longer possible as Markov-Chain Monte Carlo (MCMC) process (Metropolis-Hastings) become untenable given their size and scale \cite{ctm:blei}. \footnote{As noted earlier, simulation approaches are ideal for topic models because they are (1) theoretically backed; (2) unbiased; (3) computationally convenient \cite{ctm:blei}.} As an alternative, Blei and Lafferty introduce a fast variational inference that approximates posterior distributions for CTM. This approach underpins the computational framework for STM and STM can be thought of as identical to CTM when no metadata covariates are included.

The other two models (Dirichlet-multinomial model (DMR) and sparse additive generative model (SAGE) ) provide the framework for introducing document metadata covariates. First, the DMR model applies to the introduction of metadata covariates that can affect the topic proportions. The model replaces LDA's assumption of a Dirichlet prior for the topic distribution with a Dirichlet-Multinomial regression for the given covariates \cite{dmr:mimno}. On the other hand, researchers found that the same approach (Dirichlet-multinomial regression) was not feasible for the word distributions. Eisenstein et al. (2009) identify three main problems when applying the Dirichlet-multinomial framework to the word distribution: the increase in parameters, the computational complexity and the lack of sparsity. This is especially a problem when considering the word-topic relationship for a large corpus of documents can have an extensive vocabulary \cite{sage:eisenstein}. To address this problem, they introduce the Sparse Additive Generative Model (SAGE). The SAGE consists of an alternative framework that uses deviations in log-frequency from a benchmark distribution. There are two main advantages of this model. By cutting down on the number of parameters, this approach (1) reduces overfitting issues and (2) “can combine generative facets through simple addition in log space, avoiding the need for latent switching variables” \cite{sage:eisenstein}.

Given the inclusion of these predecessor models, it’s important to review the terminology to distinguish between the two types of covariates used in the model: prevalence and content. Prevalence covariates are document-level attributes that affect which topics are communicated in a document. In other words, prevalence covariates impact the topic proportions through the DMR model. On the other hand, a content covariate is a document-level attribute that affects how the topics are conveyed in a document. In this case, a content covariate modifies which words are used to communicate a topic and thus the word proportion that define each topic. Similarly, content covariate is used in the SAGE component of the STM model.\footnote{The current STM computation only allows one content covariate. However, this covariate can have multiple levels and thus a researcher can approximate this with multiple variables by using the interaction between all of the levels of the variables. For example, instead of having two binary content covariates for gender (male or female) and treatment (yes or no), this can be approximated with four levels of one covariate (male-yes, male-no, female-yes, female-no).}

For STM inference, like its predecessor models the exact posterior is intractable which restricts the computational methods possible. To address intractability,  Roberts et al. (2016) introduce a partially collapsed variational Expectation-Maximization algorithm for inference estimation. Another problem with the inference estimation with this model is the non-conjugacy of the logistic normal distribution, which replaces the Dirichlet prior distribution assumed in LDA, to the posterior distribution (multinomial). To account for the non-conjugacy, they also introduce a Laplace approximation for the non-conjugate elements of the model. 

\begin{figure}[!t]
\centering
\includegraphics[width=3.5in]{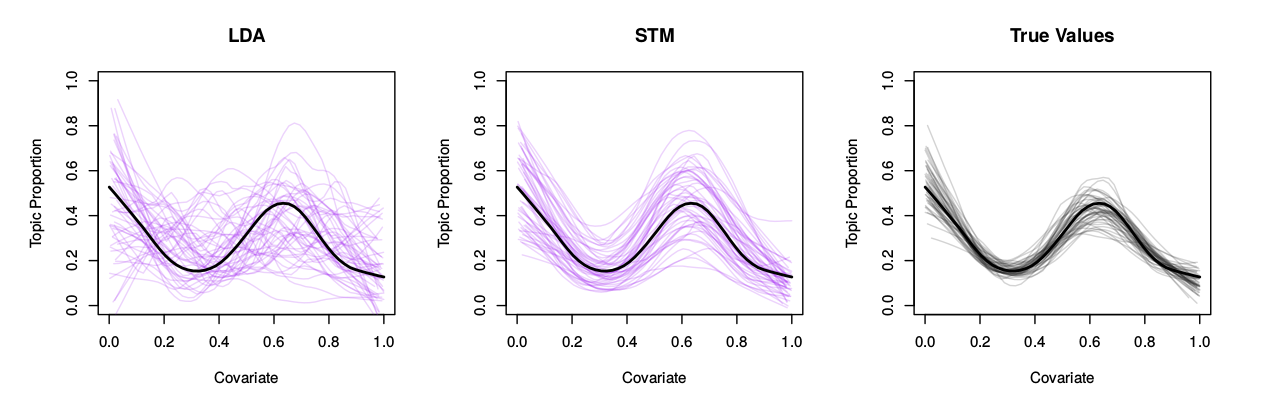}
\caption{Covariate inference for LDA and STM on simulated datasets by Roberts et al. (2016)}
\label{fig_stm}
\end{figure}

\begin{figure}[!t]
\centering
\includegraphics[width=3.5in]{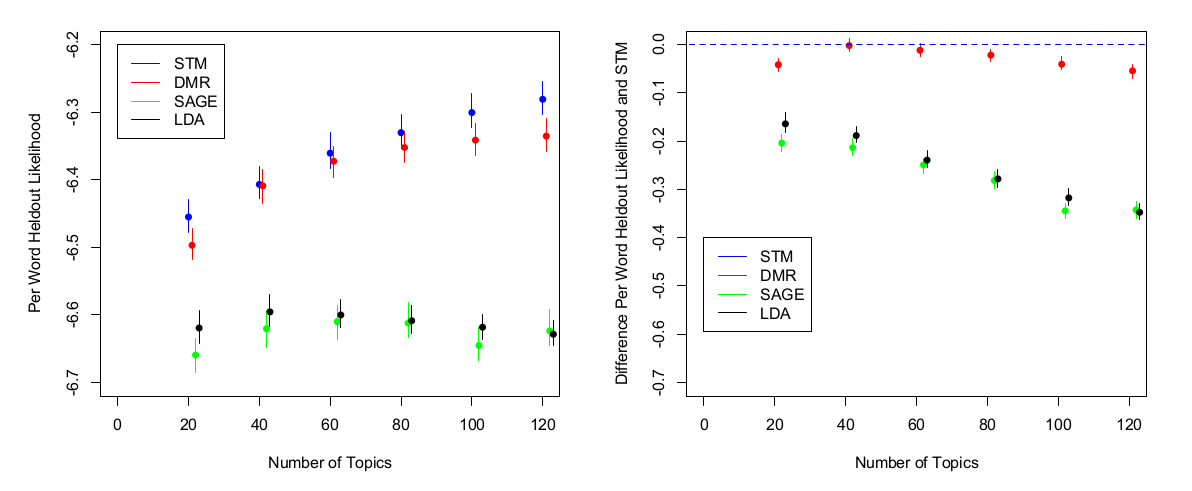}
\caption{LDA, STM, SAGE and DMR out-of-sample performance by Roberts et al. (2016)}
\label{fig_stm2}
\end{figure}

Roberts et al. (2016) analyze the estimation benefits of STM relative to LDA and STM’s sub-component models like CTM, SAGE and DMR. They find that for metadata-generated topic processes, STM outperforms LDA in covariate inference and out-of-sample prediction \cite{text:roberts}. Figure \ref{fig_stm} provides the performance of LDA and STM on 50 samples of a simulated dataset (including a random component) that is generated along with a continuous, non-linear covariate. LDA was, on average, able to identify the non-linear pattern of the covariate with the topic proportion\footnote{STM is able to model non-linear patterns through its extension to allow spline transformations and covariate interactions.}; however, due to the noise component, LDA was not robust to consistently identify the true non-linear pattern. For example, there were cases in which the model's estimates inferred a reverse pattern than the true shape value. On the other hand, STM was able to correctly identify the pattern consistently across all 50 datasets. The conclusion is by incorporating the covariates directly into the algorithm (rather than through a post-hoc analysis), we can be confident that STM will consistently detect the pattern while on average LDA can but not always. Ultimately, the improved performance of measuring covariate relationships yield the model to consistently out perform (post-hoc) LDA in out-of-sample prediction. Figure \ref{fig_stm2} is an analysis by \cite{text:roberts} on the same dataset but using LDA, SAGE, DMR and STM model separately. They find that STM, largely driven by the influence of DMR, had the best performance while SAGE and LDA performed worse across all candidate sets for a different number of topics.  

There are three major advantages to the STM model. First, the model facilitates a statistical-based framework for facilitating hypothesis testing for the causal impact of document metadata that affects which and how the topics vary by document. In fact, the model allows for general relationship framework outside of typical linear relationships in which researchers can test for non-linear patterns through log, spline or interaction terms. Second, the model introduces enhancements to the computational methods in order to make the model feasible for modeling as well as methods for model evaluation, interpretation and handling multi-modality. Last, the authors introduce an open-source R package to accompany the paper \cite{2014:roberts}. This package facilitates the implementation of the model to a much wider audience (e.g. social scientists) by providing the model in a high-level (R) rather than the traditional low-level languages most topic models methods have previously been available (e.g. Java in Mallet \cite{2002:mallet}, Python for Gensim \cite{2010:gensim}). 

\begin{figure*}[ht]
\centering
\includegraphics[width=6.5in]{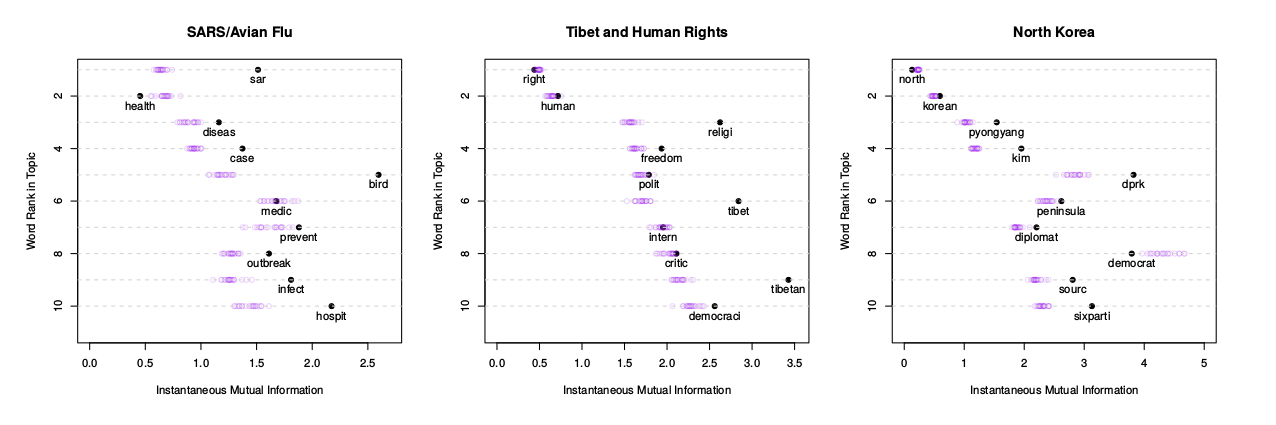}
\caption{Posterior predictive checks (PPC) using instantaneous mutual information from Roberts et al. (2016)}
\label{fig_ppc}
\end{figure*}

Despite these advancements, STM has multiple limitations. First, the model remains intractable for a large number of covariates or more than one content covariate. Second, the model can produce statistical testing for prevalent covariates but not content covariates. Third, given topic proportions' zero sum property (i.e., must sum to one), interpretation of covariates' marginal effects in STM are difficult as an increase in one topic proportion must be tied to a decrease in similar magnitude to the other topic proportions \cite{fong2016discovery}. Last, the model uses are approximation methods that are more complex than simulation-based (e.g. MCMC or Gibbs Sampling) methods. Last, as the model is more complex than LDA, so too is its output as it now must account for differences in the input covariates. With this last point in mind, I'll consider the role that explainable interactive visualizations can play in future research for STM in the conclusion of this paper.

\subsection{Multi-Modality}

The second problem with topic models is its lack of stability due to its inherent computational complexity. Topic model inference is a NP-hard problem \cite{np:sontag, localmodes:roberts}. This hardness leads to the problem of multi-modality, i.e., an optimization problem (like maximum likelihood) can be solved locally but cannot, with certainty, be solved globally.\footnote{Roberts et al. (2015) write in a footnote that connecting NP-hard complexity and multi-modality (local modes) is difficult to state easily. They argue that hardness is sufficient to prove an algorithm is not easily solved with global convergence.} Multi-modality is important when an algorithm's output can change with initialization parameters. In other words, different starting parameters can alter results and threatened the legitimacy of the approach and results.\footnote{Interestingly, Roberts et al. (2015) note that such sensitivity to starting positions is well known by computer scientists yet infrequently discussed.} While this implication cannot be directly solved (as it’s a product of the algorithm’s hardness), they argue the solution for researchers is to use improved initialization (spectral) methods and posterior predictive (stability) checks \cite{1995:gelman} to achieve the best of all possible local optima. Further, Chuang et al. (2015) provide an example of how a visualization interface can help researchers understand the effect of multi-modality on the stability of their results.

A good initialization approach needs to balance the trade-off between the computational cost of implementation and the relative model improvement for optimizing the initial state \cite{localmodes:roberts}. In the STM model, a researcher has the option to a spectral initialization that provides a quick starting point that minimizes the chance of finding sub-optimal local minima \cite{localmodes:roberts}. One approach for STM is to run standard LDA on the dataset and use the LDA results to help determine the initial state for STM. Roberts et al. (2015) find that using LDA's results for initialization not only improves the model's results but also (using Gibbs sampling) leads to a faster convergence. However, they find that using LDA for initialization does not help increase the average quality of the results with more simulations. Instead, they recommend a spectral learning approach that provides a more robust initialization with the computational complexity of using LDA's results. The spectral approach utilizes the connection of LDA with non-negative matrix factorization (see \cite{2013:arora}) that provides theoretical guarantees that the optimal parameters will be recovered \cite{localmodes:roberts}. Essentially, this approach makes stronger model assumptions (matrix decomposition elements must be non-negative) in order to avoid the problems of multi-modality. However, Roberts et al. (2015) identify two practical limitations with spectral initialization. First, it requires large amounts of data to perform adequately. Second, the modification of the model's assumptions have the potential of leading to less interpretable models. 

Posterior predictive checks (PPC) provide insight on how well the model's assumptions hold \cite{text:roberts, 2011:minmo}. Figure \ref{fig_ppc} provides one such PPC (instantaneous mutual information from \cite{2011:minmo}) for three topics, each plot representing the top ten most likely words for each of the three topics.  In these examples, the model's assumptions does not hold if there is a significant difference between what is observed (the black circle) and the simulated reference distribution (the open circles). For instance, in the ``SARS/Avian Flu'' topic, the words ``sar'' and ``bird'' are observed to be drastically outside of its reference distribution. This indicates that these words indicate would most likely be better of split between two separate topics. While one deviation like this may not jeopardize the entire topic model results, the goal of analyzing PPC is to identify systematic errors (e.g. caused by a poor local optimal solution or initialization) like this across multiple topics that can threaten the legitimacy of the model as a whole \cite{2011:minmo}. 

\begin{figure}[ht]
\centering
\includegraphics[width=3.25in]{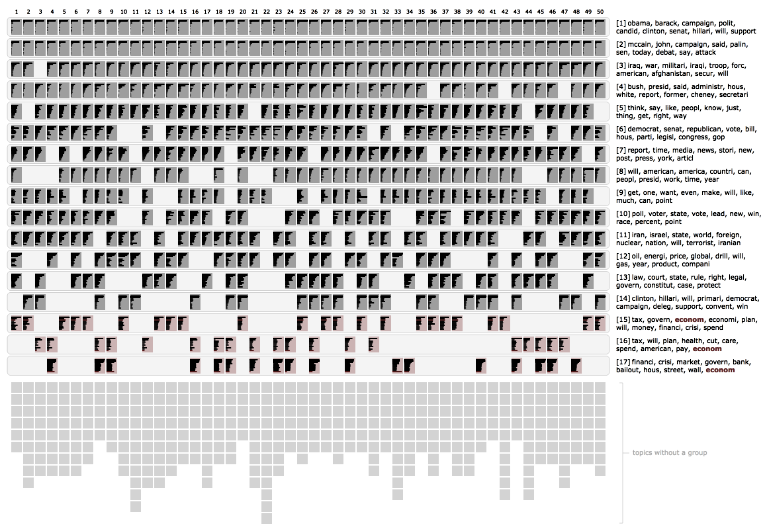}
\caption{TopicCheck with 50 Iterations of a 20 Topic STM to assess topic stability from Chuang et al. (2015)}
\label{topiccheck}
\end{figure}

Last, Chuang et al. (2015) provide an interactive solution to the problem of multi-modality: TopicCheck, a visualization interface to assess the stability of topics across multiple runs of STM \cite{2015:chuang}. Figure \ref{topiccheck} shows TopicCheck for 50 iterations of a 20 topic STM for a dataset of 13,250 political blogs. Each rectangle is a topic, with each column being one run of STM. Topics are aligned across each STM iteration into horizontal groups or rows. Topics that are not aligned to other topics (for a different STM run) are included below the baseline ``x-axis''. One of the key goals of this visualization is to determine how robust (or stable) a topic is for different STM iterations. For example, the top row (including words like ``Barack'', ``Obama'') has topics for all 50 models and thus we can feel confident this aligned topic is very robust. On the other hand, the last topic row (financial crisis) is a topic in only 18 of the 50 STM iterations. We can infer that this topic is not as stable as the topics that have aligned topics in more STM iterations. Chuang et al. (2015) findings suggest the need to consider more than one topic model as one ``single topic may not capture all perspectives on a dataset'' \cite{2015:chuang}. Further, another contribution of this paper is analyzing the impact of including or excluding rare-words on the stability of the topics. This is a novel approach to provide users an interactive understanding of the impact of a pre-processing step (like sparse word removal) that they find can potentially have a significant impact on the final topics.

\subsection{STM Applications in Social Sciences}

Already, STM has been applied to multiple areas of social science including open-ended surveys, political rhetoric, social media and massive online open courses (MOOCs). Roberts et al. (2014) provide the first extensive application of STM with open-ended survey responses.\footnote{Roberts et al. (2013) was the first paper that introduced STM and provided two simple applications (open-ended surveys and media bias in news articles). However, both of these analyses only covered a few paragraphs and did not provide the full details of the model including model setup, selection and validation.} Traditionally, human coding is the most common methodology for social scientists to analyze open-ended survey responses.\footnote{Roberts et al. (2014) acknowledge that many survey analyses simply ignore open-ended survey responses in favor of closed-ended surveys given the lack of tools to analyze such results.} The authors argue STM has two key advantages. First, being an unsupervised algorithm, STM allows the researcher to ``discover topics from the data, rather than assume them'' \cite{openended:roberts}. Second, it provides a formal way of quantifying the content and prevalance effects on the topics, especially with a treatment variable, through a cheaper, more consistent semi-automated process. 

For their application of STM, they use the ANES (American National Election Studies) survey dataset of 2,323 respondents on questions related to the 2008 election. In addition to open-ended survey responses related to the election, the dataset includes individual covariates about the respondent including their top voting issue, party identification, education and age. Further, the survey randomly assigned respondents into test and control groups in which either group was subjected to different treatment to simulate either intuitive (test) or reflective (control) thinking. The treatment variable was collected to test political science theory on the role either intuition or reflection shapes decision making as demonstrated through their open-ended responses. 

To analyze their results, the authors compare their finding with the STM model relative to that of human coders who had previously analyzed the same dataset. They make three conclusions. First, they find that in aggregate, STM categorized most of the same responses into similar topics as human coders. Second, they find that STM recovers covariate relationships closely identical to the ANES human coders. The first two findings suggest that STM can do the job very similar to the human coders but at a much lower cost through automation rather than hiring multiple human coders. Last, they do find STM has the additional advantage in that it required less assumptions about the topics themselves as STM’s unsupervised approach naturally found the occurring topics without requiring to know in advance what topics were likely to be discussed. However, they do find one disadvantage to STM is that low incidence pre-determined categories are unlikely to be identified in STM as the unsupervised approach will likely identify these topics. Instead, STM may have ``writing style'' topics using frequent words like ``go'' and ``get'' that are not relevant but occur in the unsupervised approach because they have enough volume and were not used as stop words. Nevertheless, they find that the benefits of STM largely outstrip any such costs relative to human coding in this example \cite{openended:roberts}.

Another significant application of STM has been by political scientists to analyze political rhetoric. One of the first examples come from Milner and Tingley (2015) in which they analyzed the text within lobbying reports to estimate the impact of whether the White House was lobbied on that specific issue. They find that economic interest groups tend to have higher influence in topics that have high distributional impacts and low information asymmetries like military spending. Moreover, they find in topics like military spending, such economic interest groups are less likely to directly solicit the White House, instead targeting the White House with direct lobbying efforts for efforts that are more policy focused \cite{2015:milner}.

\begin{figure*}[ht]
\centering
\includegraphics[width=6.5in]{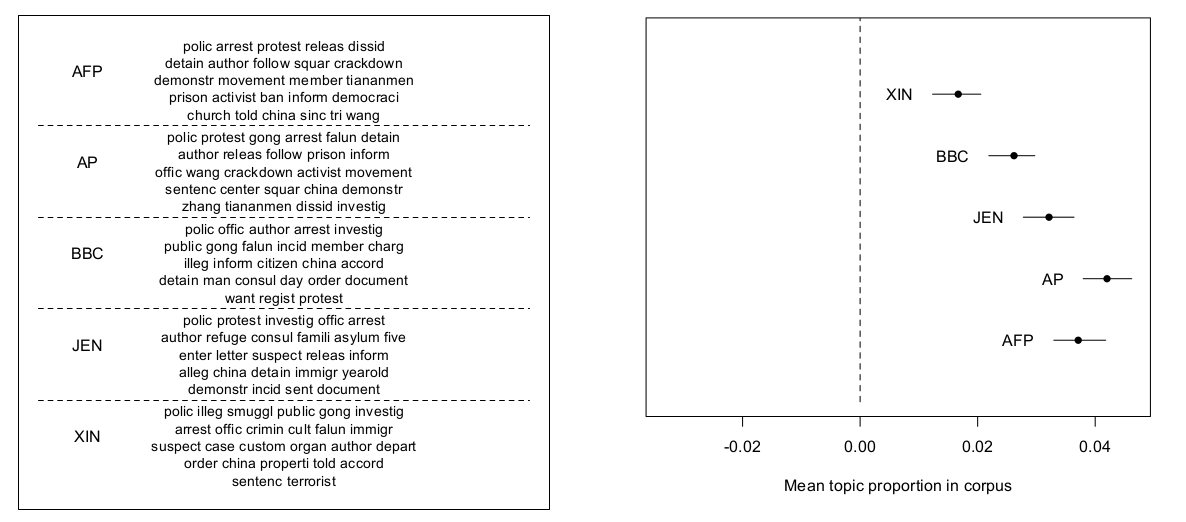}
\caption{The most likely words for a topic (``Falungong'') conditioned on the news source (Left) and the estimated topic proportion for each source (Right) from Roberts et al. (2016)}
\label{slant}
\end{figure*}

A further STM application by political scientists includes the measurement of media bias in news articles related to China by five different international media outlets.\footnote{The five media outlets are Xinhua (China), BBC (Great Britain), JEN (Japan), AP (United States) and AFP (France)} Using a dataset of over 14,000 news articles, Roberts et al. (2016) analyzed the difference in what (prevalence) and how (content) topics related to the rise of China were communicated by different media sources. They found that the Chinese-based media outlet (XIN) had, in general, a more positive view of growth-related topics than other international media sources while covering less and using different language for more controversial topics like Chinese dissidents and protesters. For example, Figure \ref{slant} provides two graphics relating the topic relating to the Falungong movement, a Chinese religious group that the Chinese government outlawed in 1999 leading to many protests and a government crackdown on participants. The left side represents the different words used to characterize this topic conditioned on each media source (content covariate). Xinhua (XIN) used words like ``illegal'', ``smuggler'', and ``criminal'' to describe the movement as an illegal operation. On the other, Western media outlets like the U.S. Associated Press (AP) and the French Agence France-Presse (AFP) characterized the movement with language characterizing the movement as a protest rather than a criminal operation. For example, AP and AFP's conditioned topic words include ``protest'', ``dissident'', ``movement'' and ``crackdown''. Moreover, STM could also be used to measure the difference in topic proportion each outlet used each topic. In the example of the Falungong topic, the right side of Figure \ref{slant} shows the estimated topic proportions (along with confidence intervals) for each of the five media outlets. Xinhua covered the Falungong topic much less frequently than did Western media outlets like AP and AFP with statistical significance. 

In addition to political and news rhetoric, other applications of STM by social scientists include social media messages (in particular Twitter) and online course feedback. Lucas et al. (2015) analyzed bilingual social media messages through automated machine translation to estimate the effect language (Arabic or Chinese) has on Twitter users' topics regarding Edward Snowden \cite{2015:lucas}. Sachdeva et al. (2016) used STM to analyze smoke-related tweets and the potential spatial-temporal effects of wildfires have on users' tweets relative to those individuals who reside or work close to affected areas \cite{2016:sachdeva}. Reich et al. (2015) analyzed the use of massive online (MOOC) course feedback to scale open-ended course responses. Reich et al. (2016) extended this work in MOOCs and connected it with political rhetoric to identify political discussions within these courses. Ultimately, given STM's recent introduction, these examples represent just a handful of the early applications. As the technique begins to mature and more researchers learn about the methodology, there is no doubt the number and the range of applications will begin to increase rapidly in the near future.

\section{Conclusion and Future Research}

While used rarely by most social scientists, topic models offer social scientists an innovative and objective way to measure latent qualities on large, unstructured datasets like social media, open-ended surveys and news or research publications. Topic models are one of many machine learning frameworks that, when combined with causal inference tools, represent a significant opportunity for social scientists to answer many of society's large-scale problems \cite{grimmer2015we}. In particular, STM represents an example of integrating machine learning (topic models) with causal inference mechanisms in a generalized framework that can be applied to many social science problems. Nevertheless, a major impediment to the expansion of STM (and machine learning algorithms in general) for most social scientists is the high knowledge barriers to use these models. Given the quick development of such models, many social scientists may lack the training and experience in machine learning and computational programming to implement and to analyze topic models. To address this concern, a potential research opportunity with the STM model is through an explainable user interface (visualization) that can aid social scientists in hypothesis testing large collections of text documents. The importance of this opportunity is exemplified through DARPA's recent Explainable Artificial Intelligence (XAI) program \cite{2016:darpa}. 

\begin{figure}[h!]
\centering
\includegraphics[width=3.5in]{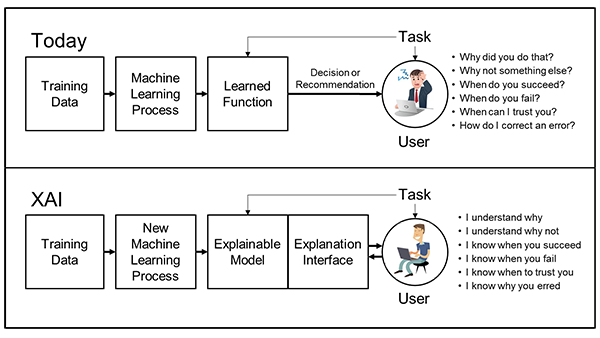}
\caption{Explainable Artificial Intelligence (XAI) from DARPA (2016)}
\label{fig_darpa}
\end{figure}

An inherent problem with most machine learning techniques and tasks for decision makers is the lack of explanation of why a specific prediction or choice was made by the algorithm. Instead, the goal of the XAI program is to provide explainable features to traditional machine learning techniques along with an explainable user interface (e.g., visualization) to combine human insight with the predictions of the model to answer the question of why. Structural topic modeling would directly fit under the program's approach of interpretable models that ``that learn more structured, interpretable, or causal models'' \cite{2016:darpa}.\footnote{In addition to interpretable models, the program cites two other approaches: deep explanation and model induction.} Figure \ref{fig_darpa} provides a graphic representing the role of explainable user interfaces for machine learning tasks like STM inference.

Therefore, a key research opportunity for STM is the development of an explainable, intelligent interactive system \cite{2015:kulesza} to analyze for interpretation, model evaluation, multi-modality, pre-processing steps\footnote{Interactive interfaces could be used to measure the impact of pre-processing (stemming, stop words, etc.) to help researchers determine its impact on the model's results.}  and validation. Such an interface could be built integrating high level visualization tools like Shiny \cite{shiny} and Vega-lite \cite{2015:wongsuphasawat}, \cite{2017:stayanarayan} that can provide a robust set of tools with minimal amount of coding. Further, if written in R, such an interface could easily combine with R packages like stm for widespread use (see Appendix 1).\footnote{Currently, there are two STM-based visualization packages (stmViz and stmBrowser) but their functions are limited.}


%

\appendices
\section{Appendix 1: stm R package}

\begin{figure}[ht]
\centering
\includegraphics[width=3.25in]{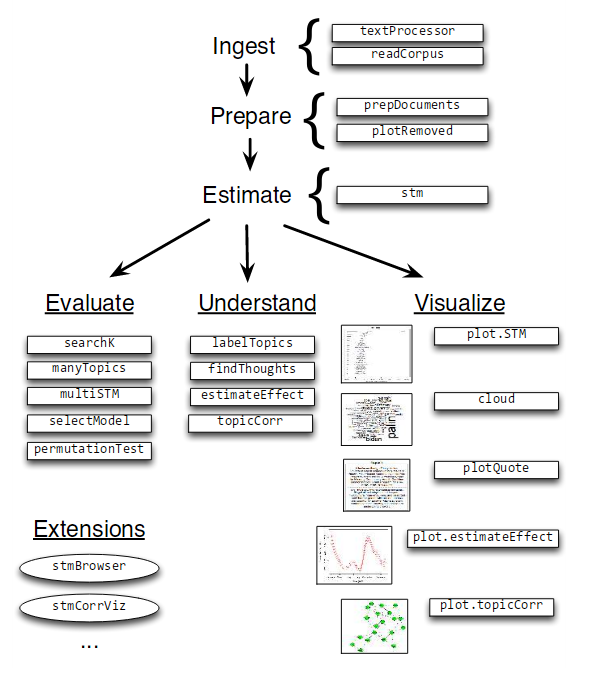}
\caption{Functions within stm package from Roberts et al. (2015b)}
\label{fig_rpack}
\end{figure}

The R package stm was introduced by Roberts et al. (2014b) to facilitate the widespread use of STM for R users \cite{2014:roberts}. A key advantage to the stm R package is that includes multiple methods of posterior predictive checks, interpretation, data preprocessing, model selection and static visualizations. Figure \ref{fig_rpack} provides an outline of the functions within the stm package categorized by their different functionalities.

\section{Appendix 2: Word Embedding Models - word2vec and GloVe models}

More generally, topic models are part of a larger framework of language models called vector space models. In this approach, language is encoded by vector representations based on the Distributional Hypothesis of language, i.e., words are used in similar context to words with similar meanings \cite{baroni2014don}. Under this interpretation, bag-of-words models (like LSA and LDA) are considered as count-based models as they encode language as a series of vector counts. As mentioned earlier, one downside of this approach is that it ignores the context in which words are used, an obvious deficiency when considering semantic meaning. Alternatively, context-based models are a new generation of vector space models that encode word context within a vector framework. These models are sometimes called word embedding, neural language models or simply predictive models. 

The first example of context based models was introduced by Mikolov et al. \cite{mikolov2013efficient} as the word2vec models: Continuous Bag of Words (CBOW) and Skip grams. Unlike the traditional bag-of-words approaches that treated each word like an atomic unit, this approach considered a rolling window context (e.g. five words) to build a vector space model that provided deeper semantic meaning while facilitating scalability to millions of words and documents. The difference between the models is a reversal of the model inputs and outputs. The CBOW model uses a list of words (e.g. five words) to best predict a word that will most likely be used in a similar context. Alternatively, the skip gram model uses a given word to predict what are the most likely words that will be used in a similar context. 

Building off of this framework, Pennington, Socher and Manning \cite{pennington2014glove} unified vector space models by combining features of the count-based models (like LSA and LDA) and context-based models (like word2vec) to a more robust model named GloVe, or global vectors of word representations. Acknowledging the deficiencies of the count-based models that motivated the word2vec model, Pennington, Socher and Manning argue that word2vec models have the opposite problem that by only analyzing the (local) context of words, this approach fails to utilize the statistical properties provided through a count-based approach. Ultimately, their approach resulted in not only a better predictive model that produced deeper semantic meanings, the neural language structure \cite{bengio2003neural} connected directly with related breakthroughs in the application of deep learning to text analysis \cite{lecun2015deep}.



\ifCLASSOPTIONcaptionsoff
  \newpage
\fi



%

%

\begin{IEEEbiography}[{\includegraphics[width=1in,height=1.25in,clip,keepaspectratio]{picture}}]{John Doe}
\blindtext
\end{IEEEbiography}




\end{document}